\pdfoutput=1
\documentclass[11pt]{article}

\usepackage{authblk}

\usepackage{emnlp2021}

\usepackage{times}
\usepackage{latexsym}
\usepackage{amsmath}

\usepackage[T1]{fontenc}
\usepackage[utf8]{inputenc}

\usepackage{microtype}

\usepackage{booktabs}

\usepackage{textcomp}
\usepackage{graphicx}
\usepackage{multirow} 
\usepackage{enumitem}
\usepackage{arydshln}
\newcommand{\tldrl}{\textsc{Tldr9+}}
\newcommand{\tldrs}{\textsc{TldrHQ}}
\newcommand{\tldr}{\textsc{Tldr}}
\usepackage{etoolbox}
\makeatletter
\patchcmd{\maketitle}
 {\def\@makefnmark}
 {\def\@makefnmark{}\def\useless@macro}
 {}{}
\makeatother

\title{\tldrl: A Large Scale Resource for \\Extreme Summarization of Social Media Posts}

\author[1*]{\textbf{Sajad Sotudeh} \thanks{*Work done during the internship at Adobe Research.}}
\author[2]{\textbf{Hanieh Deilamsalehy}}
\author[2]{\textbf{Franck Dernoncourt}}

\author[1]{\textbf{Nazli Goharian}}

\affil[1]{IRLab, Georgetown University}
\affil[ ]{{\fontfamily{lmtt}\selectfont{ \{sajad,nazli\}@ir.cs.georgetown.edu}}}

\affil[2]{Adobe Research} 
\affil[ ]{{\fontfamily{lmtt}\selectfont{
\{deilamsa,franck.dernoncourt\}@adobe.com}
}}

\begin{document}
\maketitle
\begin{abstract}
Recent models in developing summarization systems consist of millions of parameters and the model performance is highly dependent on the abundance of training data. While most existing summarization corpora contain data in the order of thousands to one million, generation of large-scale summarization datasets in order of couple of millions is yet to be explored. Practically, more data is better at generalizing the training patterns to unseen data. In this paper, we introduce \tldrl{} ---a large-scale summarization dataset--- containing over 9 million training instances extracted from Reddit discussion forum (\url{https://github.com/sajastu/reddit_collector}). This dataset is specifically gathered to perform \textit{extreme} summarization (i.e., generating one-sentence summary in high compression and abstraction) and is more than twice larger than the previously proposed dataset. We go one step further and with the help of human annotations, we distill a more fine-grained dataset by sampling \textbf{H}igh-\textbf{Q}uality instances from \tldrl{} and call it \tldrs{} dataset. We further pinpoint different state-of-the-art summarization models on our proposed datasets.

\end{abstract}

\section{Introduction}
Text summarization is defined as generating a concise sequence of text as summary, given relatively a longer document as source. A high-quality summary conveys the most important points of its associated source. The task is generally performed in two ways: 1) extractive in which salient sentences are identified and concatenated to form the final summary~\cite{Nallapati2017SummaRuNNerAR, Dong2018BanditSumES,Sotudeh2020OnGE, Narayan2020StepwiseES, Cho2020BetterHC}; and 2) abstractive that produces a paraphrasing of the main contents of the given text.~\cite{See2017GetTT,Gehrmann2018BottomUpAS, MacAvaney2019OntologyAwareCA, zhang2019pegasus, Sotudeh2020AttendTM, Lewis2020BARTDS, Lebanoff2020LearningTF} and is considered more challenging as the model needs to deal with novel words generation beyond sentence extraction.

Over the past few years, different neural models including RNN~\cite{Hochreiter1997LongSM} and Transformer-based~\cite{Vaswani2017AttentionIA} networks have been proposed to facilitate the summarization task. While promising, the performance of such models is bound to the abundance of training data due to the massive model complexity~\cite{Ying_2019}. Lack of sufficient training data worsens the model's ability to generalize patterns in training data to unseen data~\cite{Althnian2021ImpactOD}. In addition, overfitting will be likely inevitable as the model is forced to learn from a limited set of data; hence, hindering the generalization. This justifies the necessity of large-scale corpora for training large and complex models. 

\begin{figure}
    \centering
    \includegraphics[scale=0.57]{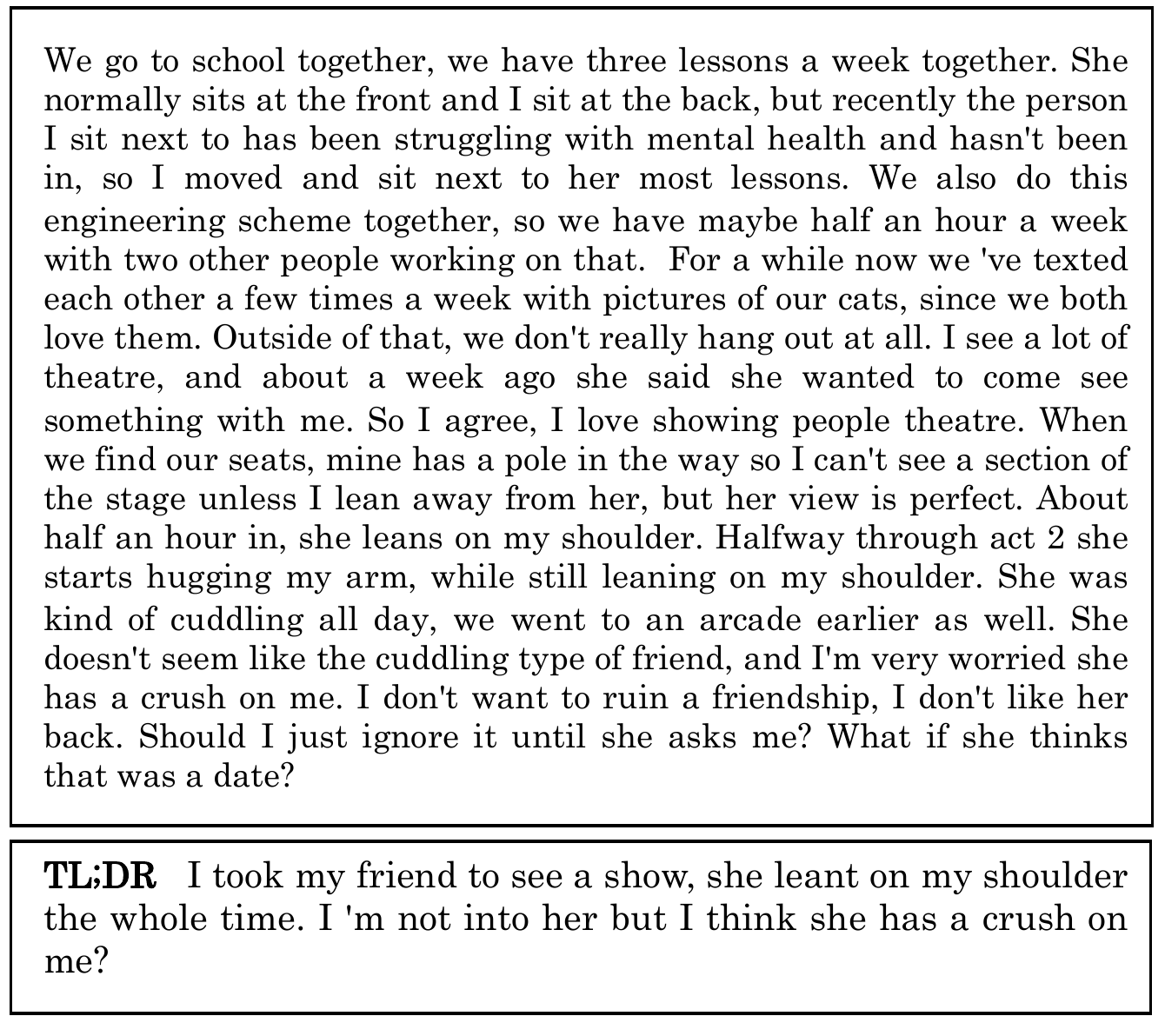}
    \caption{An example Reddit post with \tldr{} summary. As seen, the \tldr{} summary is extremely short, and highly abstractive.}
    \label{fig:tldr_example}
\end{figure}

Prevalence of social media platforms has provided communities with an opportunity to exchange different types of data while interacting with each other. \textit{Reddit}~\footnote{\url{https://www.reddit.com/}} is one of such popular platforms where users post their content of interest in a variety of domains. \tldr , the acronym for ``Too Long; Didn't Read'', is a common practice that aims at removing unnecessary information from the lengthy post, and presenting its gist information in a few words. Figure \ref{fig:tldr_example} shows a sample of Reddit post with its \tldr , which aims at abstracting post with extreme compression. Abundance of posts that contain such \tldr s during recent years has given rise to generation of data collections that can be utilized for training deep neural networks; hence, addressing the challenge of large-scale datasets' scarcity. Despite the possibility of acquiring large-scale datasets from social media platforms, training deep neural networks on such datasets is yet challenging. This might be due to the specific writing style of social media content such as \textit{informal} language and massive \textit{noise} within such content~\cite{Sotudeh2020GUIRAS}.

\begin{table}[t]
    \centering
    \resizebox{\columnwidth}{!}{%
         \begin{tabular}{lrlr}
         \toprule
         \textbf{Dataset} & &  \textbf{Domain} & \textbf{\# instances} \\
         \midrule
        %  \multicolumn{4}{l}{\textit{\small (a) Large-scale datasets}} 
        %   \\\hdashline \vspace{-0.3cm}\\
         
        %  Gigaword && News & 4M \\ 
        %  NYTimes & & News & 655K \\
        % BigPatent && Legal & 1.3M \\
        %  Newsroom & & News & 1.32M \\
        %  XSum & & News & 227K \\
        %  arXiv & & Scientific & 215K \\
        % \midrule
        %  \multicolumn{4}{l}{\textit{\small (b) Extreme summarization datasets}} 
        %   \\\hdashline \vspace{-0.3cm}\\
        \multicolumn{4}{l}{\textit{\small Non-social media}} 
          \\\hdashline \vspace{-0.3cm}\\
         \textsc{SciTldr} && Scientific & 3.2K \\
          XSUM & & News & 227K \\
          \midrule
          
          \multicolumn{4}{l}{\textit{\small Social media}} 
          \\\hdashline \vspace{-0.3cm}\\
         Reddit TIFU & & Social Media& 120K \\
          Webis-TLDR-17 & & Social Media & 4M    \\
          \hdashline  
          
      \textbf{\tldrs{} (ours)} && {Social Media} & \textbf{1.7M} \\
       \textbf{\tldrl{} (ours)} && {Social Media} & \textbf{9.2M} \\
         \bottomrule
    \end{tabular}
    }
    \caption{Overview of \textit{extreme} summarization datasets across different social and non-social domains with number of instances.}
    \label{tab:ds_stat}
\end{table}

Table \ref{tab:ds_stat} shows some of the existing summarization datasets in social and non-social media domains. These datasets are specifically proposed for \textit{extreme summarization} task, where the aim is to produce one to two summary sentences in extreme compression and high abstraction. 
% While being large-scale to some extent, none of those cross over the limitation of 4 millions instances. Hence,
In this paper, we introduce our dataset, \tldrl{} with over 9 million instances which is more than twice larger than the previous dataset~\cite{Vlske2017TLDRMR}.  We further sample high-quality instances in virtue of human annotations from \tldrl{} to construct \tldrs{} yielding 1.7 million instances in the hope of providing firm grounds for future work. Owing to extremely short length of \tldr{} summaries (less than 40 words), our datasets are rather suitable for \textit{extreme summarization} task than for longer ones.

\begin{figure*}
    \centering
    \footnotesize
\scalebox{0.98}{
    \begin{tabular}{cc}
         \multicolumn{1}{c}{\includegraphics[scale=0.55]{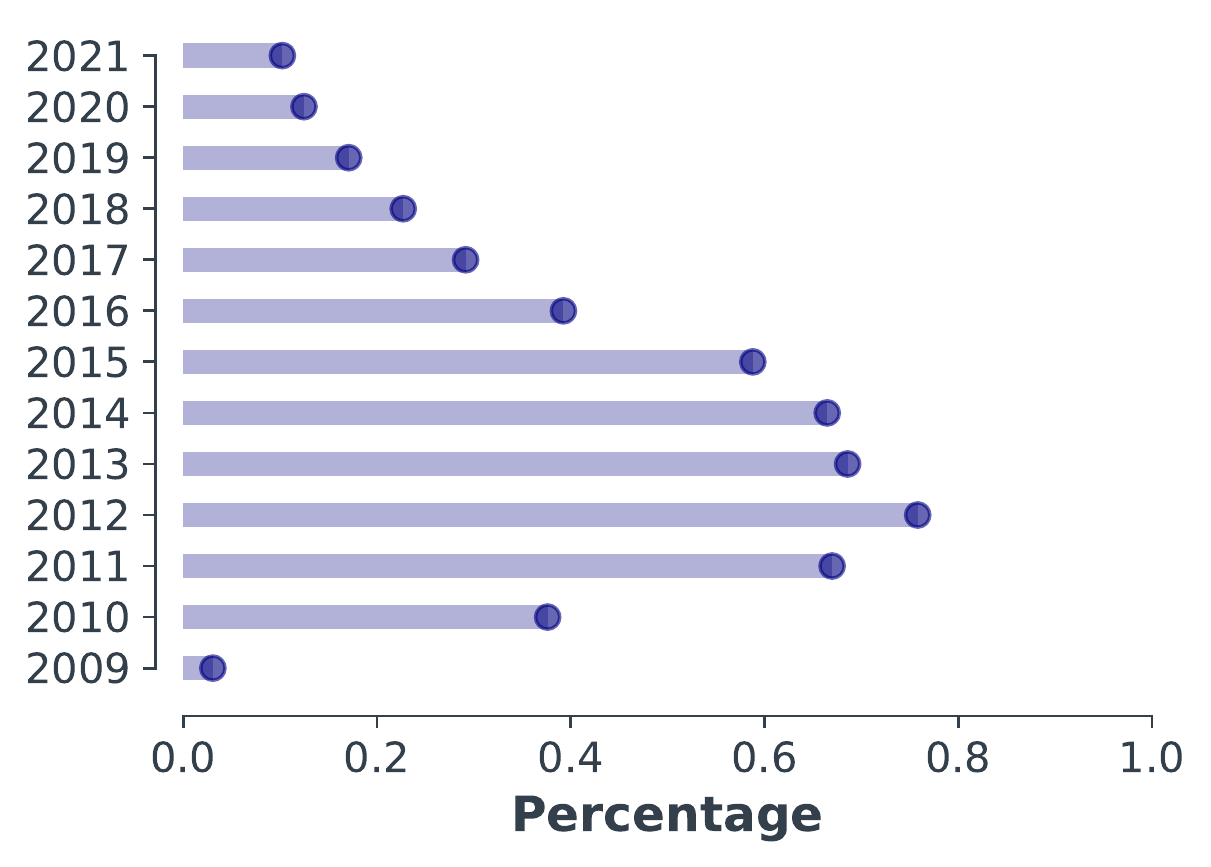}} & \multicolumn{1}{c}{\includegraphics[scale=0.55]{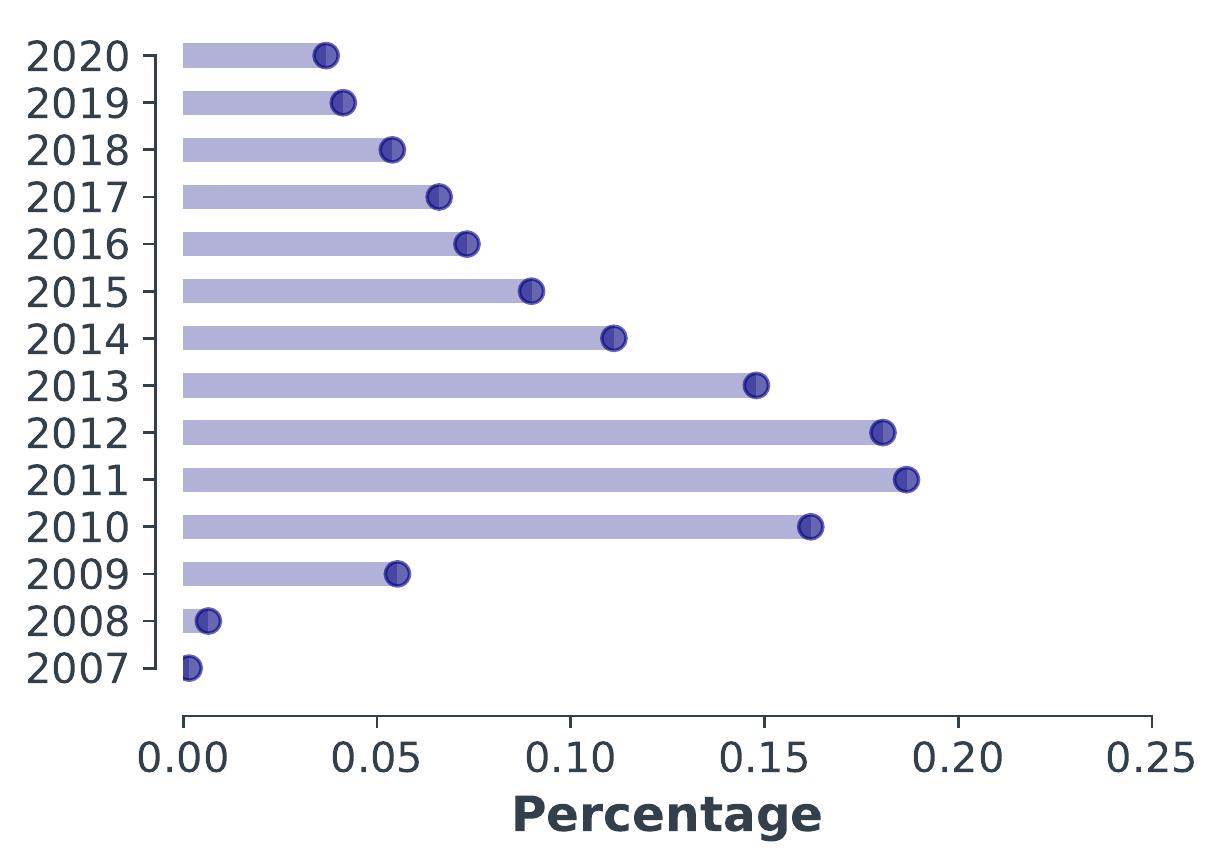}} \\
         \multicolumn{1}{c}{(a) submission-\tldr} & \multicolumn{1}{c}{(m) comment-\tldr}  \\
         
        %  \multicolumn{1}{l}{\includegraphics[scale=0.18]{figures/RS2020.pdf}} & \multicolumn{1}{l}{\includegraphics[scale=0.18]{figures/RS2020.pdf}} \\
        %  \multicolumn{1}{c}{(c) submission-TLDR} & \multicolumn{1}{c}{(d) comment-TLDR}\\
        
        %   \\
        %  \includegraphics[scale=0.52]{samples/figs/figanal2.pdf}  \\
        %   \\
         
        %  (b) Sentence graph visualization of our system's generated summary. Green and gray nodes are introductory and non-introductory sentences, respectively. Edge thickness denote the \textsc{Rouge} score strength between pair of sentences.\\
        \label{fig:tldr_proportion}
    \end{tabular}
}
    
    \caption{The proportion of \tldr s over entire posts (submissions and comments) submitted per year (Figures (c) and (d)). At the time of writing this paper, submissions dumps are partly uploaded for 2021 (until 2021-06), while there is no comments dumps uploaded for 2021.}
    \label{fjg:tldrPerc}
\end{figure*}

In this research, we aim at harvesting instances that include \textsc{Tldr}s written by the Reddit users spanning the period of 2005-2021.  Our early attempt at gathering such instances yields over 9 million instances with \textsc{Tldr}s as the initial set (i.e., \tldrl). Since social media posts are inherently noisy, we consider applying a heuristic method to cut out low-quality instances from the initial set, which ultimately results in 1.7 million high-quality instances. For deciding such heuristic, we employ human annotators to help to obtain a more fine-grained dataset (i.e., \tldrs{}). Furthermore, we establish various state-of-the-art extractive and abstractive summarization models on our proposed datasets. Finally, we carry out an analysis over the results on both datasets to shed light on future direction.  We believe that our datasets can be utilized to pave the path for future research. Our miner code and data are made publicly available at \url{https://github.com/sajastu/reddit_collector}, along with the licensing details included.

\section{Related work}
Over the past few years, summarization community has witnessed a variety of summarization datasets in different domains~\cite{See2017GetTT, Cohan2018ADA, Kornilova2019BillSumAC, Grusky2018NewsroomAD, Sotudeh2021OnGE}. While these collections have provided a fair basis to perform different neural text summarization models, the necessity of introducing large-scale collections, in magnitude of over 4 million, has not been much explored.
% Table \ref{tab:ds_stat} presents some of those large-scale datasets in various domains. 

Among the first attempts on this track, \citet{Rush2015ANA} gathered the English Gigaword corpus~\cite{graff2003english} which contains around 4 million article-headline pairs for the task of news headline generation. Researchers have noted that \textit{lead bias} is the common phenomenon in most news datasets, where early parts of the article generally include the most important information~\cite{Kedzie2018ContentSI, Zhu2019MakeLB, Grenander2019CounteringTE}. To alleviate the lead bias for training summarization models, there have been recent efforts to propose summarization datasets, where the lead bias phenomenon is mitigated and summaries are sampled from diverse source regions. Amongst those, ~\citet{Sharma2019BIGPATENTAL} proposed \textsc{BigPatent}, consisting 1.3 million patent documents, collected from Google Patents Public Datasets, with human-written abstractive summaries. \citet{Kim2019AbstractiveSO} proposed \textit{Reddit TIFU} in which the abstractive gold summaries are sampled from diverse regions of the source document, rather than lead regions. 

Our proposed datasets are more suited for the task of \textit{extreme summarization}~\cite{Narayan2018DontGM, Cachola2020TLDRES}, where the task is to create a short one-sentence summary. To this end, \citet{Narayan2018DontGM} proposed \textit{XSUM} dataset which is a real-word dataset compiling online articles from the British Broadcasting Corporation (BBC). \textsc{TLDR} generation task is also a new form of extreme summarization. \citet{Kim2019AbstractiveSO} collected \textit{Reddit-TIFU} dataset, consisting of 120K posts from the online discussions from Reddit. Recent efforts have mined around 4 million Reddit posts along with their \textsc{Tldr} summaries~\cite{Vlske2017TLDRMR} which resulted in Webis-TLDR-17 dataset. While our work is similar to theirs, our collected dataset is more than twice larger than the one previously proposed.

\section{The \textit{Reddit} Collection}
\subsection{Data Collection}
Reddit is a social news aggregation and discussion website platform that has been officially launched since June 2005. It supports some features specific to social platforms such as web content rating through up-voting, and discussion topics via subreddits. The user-created content can be of any domain such as News, Politics, Science, Sport and etc. Users can post or comment on a specific topic that falls into a specific subreddit. Within subreddits, users submit their post as \textit{submission}, and others can react through commenting under the posted submission. Each submission and comment has a \textit{text body/selftext} which reflects the users' information exchange regarding a specific topic. The existence of social platforms such as Reddit has provided the research community with an opportunity to experiment with resources that use \textit{informal} language, rather than those in news, scientific or legal documents which use formal language. 

\tldr ---Too Long; Didn't Read--- is a common practice in Reddit that often appears at the end of long Reddit posts. It is denoted as an extremely short summary that urges users to read a shorter version of a longer text when they do not have time to read the entire post. Figure \ref{fjg:tldrPerc} shows the ratio of posts containing such \tldr{} summaries over the entire submitted posts (and comments) across different years. It is observable that although we see an ascending trend since 2005, the number of \tldr s remains fixed (see Section \ref{sec:dataset_analysis}) while the number of posts increases drastically.

Pushshift~\footnote{\url{https://files.pushshift.io/}} is a social media data repository platform that has been recently made available to NLP researchers~\cite{Baumgartner2020ThePR}. It contains recent and historical dumps of Reddit posts that are updated in real-time. In order to create the \tldr{} dataset, we downloaded the whole data dumps (submissions and comments) which cover the period of 2005-2021, and extracted instances that contain \tldr s within the posted source text. This mining process resulted in \tldrl{} dataset, which contains over 9 million instances. To acquire a more fine-grained dataset, with the help of human annotations, we obtained \tldrs{} dataset, consisting of 1.7 million high-quality instances. The datasets' construction details are discussed in what follows.

% \begin{table}[h]
%     \centering
%     \begin{tabular}{lr}
%         \toprule
%          Keyword& Frequency \\
%         \midrule
%         TLDR & 12345 \\
%         \bottomrule
%     \end{tabular}
%     \caption{Caption}
%     \label{tab:my_label}
% \end{table}

\subsection{Datasets Construction: \tldrl{} and \tldrs{}}
\noindent \textbf{\tldrl.}  After downloading Reddit data dumps, we extract posts in which a mention of \textsc{Tldr}-style keywords is found. To find \textsc{Tldr}-style keywords within a given text, we declare a regular expression that matches words starting with ``TL'' and ending with ``DR'', with permission of having up to three characters in-between as also done by \citet{Vlske2017TLDRMR}. This stage yields the \textbf{\tldrl} dataset as the \textit{full} corpus. At the next filtering stage, we utilize a heuristic method along with human supervision to narrow it down to a more fine-grained dataset that contain high-quality instances. 
%  \red{Table \ref{tab:tldr_keywords}} shows top 10 TLDR keywords that are used in comments and submissions with number of occurrences. It is interesting that since choosing TLDR keywords is not conventional among authors, the authors tend to use diverse keywords when writing a TLDR. 

\noindent \textbf{\tldrs.} A few studies have noted that user-generated content in social media platforms is noisy~\cite{Liu2015EstimatingUL} in terms of having spam, bad grammar, and spelling errors. To filter out such noisy instances from the \tldrl{} dataset, we use a heuristic method to drop low-quality instances while retaining high-quality ones. To be more specific, given a post-\tldr{} pair, we firstly identify the highest score source sentence in terms of \textsc{Rouge-2} and \textsc{Rouge-L} mean scores (i.e., \textit{oracle} sentence). The choice of oracle sentence lies in the fact that we postulate to extract a sentence from the longer post that has the highest similarity with the \tldr{} summary as the gold standard. We then decide to either drop or retain the instance if the score surpasses a pre-defined \textit{threshold}. We experiment with different thresholds of 0.15, 0.17, 0.20, 0.22 and 0.25, and choose one considering the annotations done by human annotators. The details of human annotation process is discussed in what follows. 

\subsection{Human Annotation}
As mentioned earlier, we first define 5 fixed thresholds including 0.15, 0.17, 0.20, 0.22, and 0.25 to create 5 data subsets from \tldrl{} dataset. Specifically, we take \tldrl{} as the initial seed, from which 5 subsets is created as follows. To gather instances for each of the pre-defined thresholds, we check if the oracle sentence's score in the given instance surpasses the experimented threshold. If it does so, we add it to the subset, otherwise, it is dropped. We then randomly sample 20 cases from each of these subsets with their oracle sentence and \textsc{Tldr} summaries, yielding 100 cases for annotation in total. We have four human annotators from our NLP group either confirm (labeling with 1) or reject (labeling with 0) if the oracle sentence \textit{validates} the \textsc{Tldr} summary. By definition, the sentence \textit{validates} the \tldr{} summary if at least one fragment can be found within the sentence that semantically occurs in \tldr{} summary. 

We further provide the instances' text (i.e., source) as the ``Context'' for the oracle sentence, and ask the annotators to confirm or reject if the context also validates the \tldr{} summary. Context is specifically important for the cases where the oracle sentence does not validate the \textsc{Tldr} summary. In fact, by providing context, we aspire to verify if an ideal summarizer is able to generate the \textsc{Tldr} using the context when the oracle sentence is not much informative. For tie cases~\footnote{Suppose a case where two annotators confirm (label 1), while the other two reject (label 0).}, we employ a fifth annotator to make the final decision. 
% To clarify, Figure \ref{fig:annotation_samples} shows some samples along with their assigned labels by the annotators.
 \begin{figure*}[t]
    \centering
    \footnotesize
\scalebox{0.98}{
    \begin{tabular}{ll}
         \multicolumn{1}{c}{\includegraphics[scale=0.4]{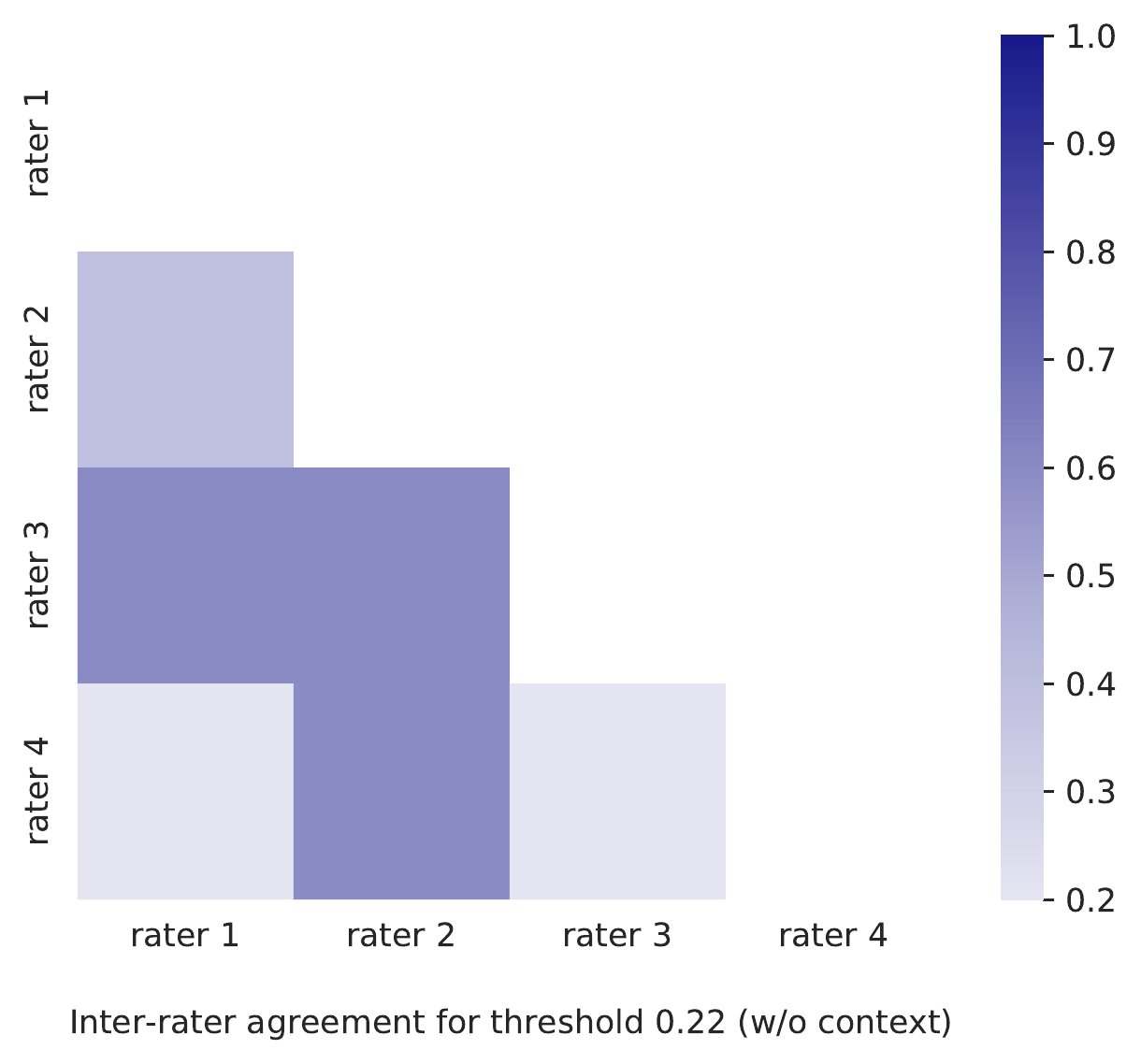}} &  \multicolumn{1}{c}{\includegraphics[scale=0.4]{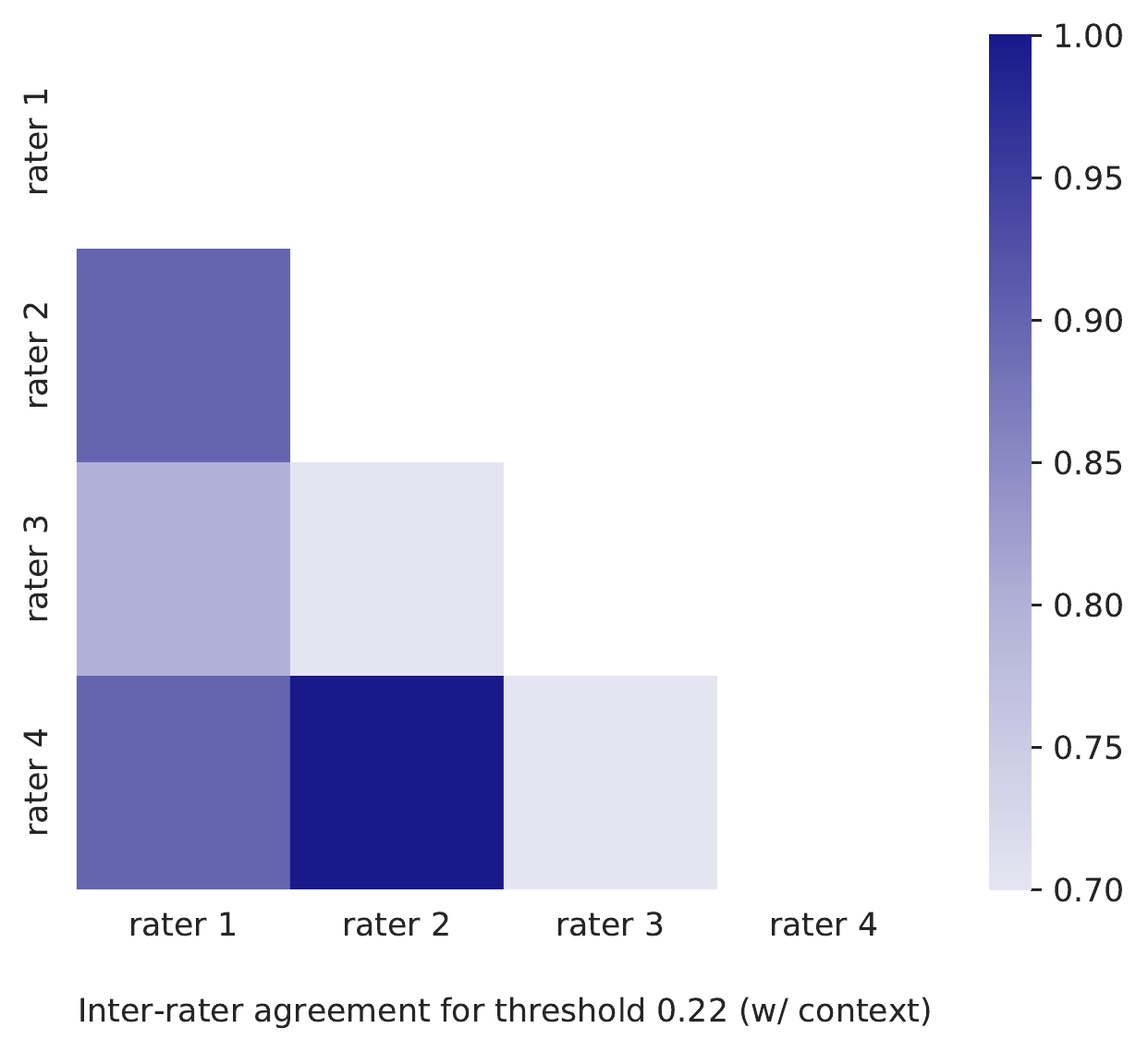}} \\
         \multicolumn{1}{c}{(a) Annotation with the oracle sentence} & \multicolumn{1}{c}{(b) Annotation with context} 
        %  \multicolumn{1}{l}{\includegraphics[scale=0.18]{figures/RS2020.pdf}} & \multicolumn{1}{l}{\includegraphics[scale=0.18]{figures/RS2020.pdf}} \\
        %  \multicolumn{1}{c}{(c) submission-TLDR} & \multicolumn{1}{c}{(d) comment-TLDR}\\
        
        %   \\
        %  \includegraphics[scale=0.52]{samples/figs/figanal2.pdf}  \\
        %   \\
         
        %  (b) Sentence graph visualization of our system's generated summary. Green and gray nodes are introductory and non-introductory sentences, respectively. Edge thickness denote the \textsc{Rouge} score strength between pair of sentences.\\
        \label{fig:tldr_proportion}
    \end{tabular}
}
    \caption{$S$ score inter-rater agreement for annotation without context (left), and annotation with context (right)}
    \label{fig:agreement_rate}
\end{figure*}
\begin{table}[h]
    \centering
    \resizebox{\columnwidth}{!}{%
    \begin{tabular}{lrrrr}
    \toprule
        Threshold  &  score w/o context & score w/ context \\
        % & & (words/sent.) & (words/sent.)& ratio \\ 
         \midrule
         0.15  & 0.65 & 0.90   \\
         0.17  & 0.90 & \textbf{1.0}   \\
         0.20  & 0.85 & 0.95   \\
         0.22  &\textbf{ 1.0} &\textbf{ 1.0}   \\
         0.25  & 0.75 & 0.90   \\
         \bottomrule
    \end{tabular}
    }
    \caption{Average decision scores given by the annotators for each threshold.}
    \label{tab:ann_score}
\end{table}

\begin{figure*}[t]
    \centering
    \footnotesize
\scalebox{0.9}{
    \begin{tabular}{cc}
         \multicolumn{1}{c}{\includegraphics[scale=0.19]{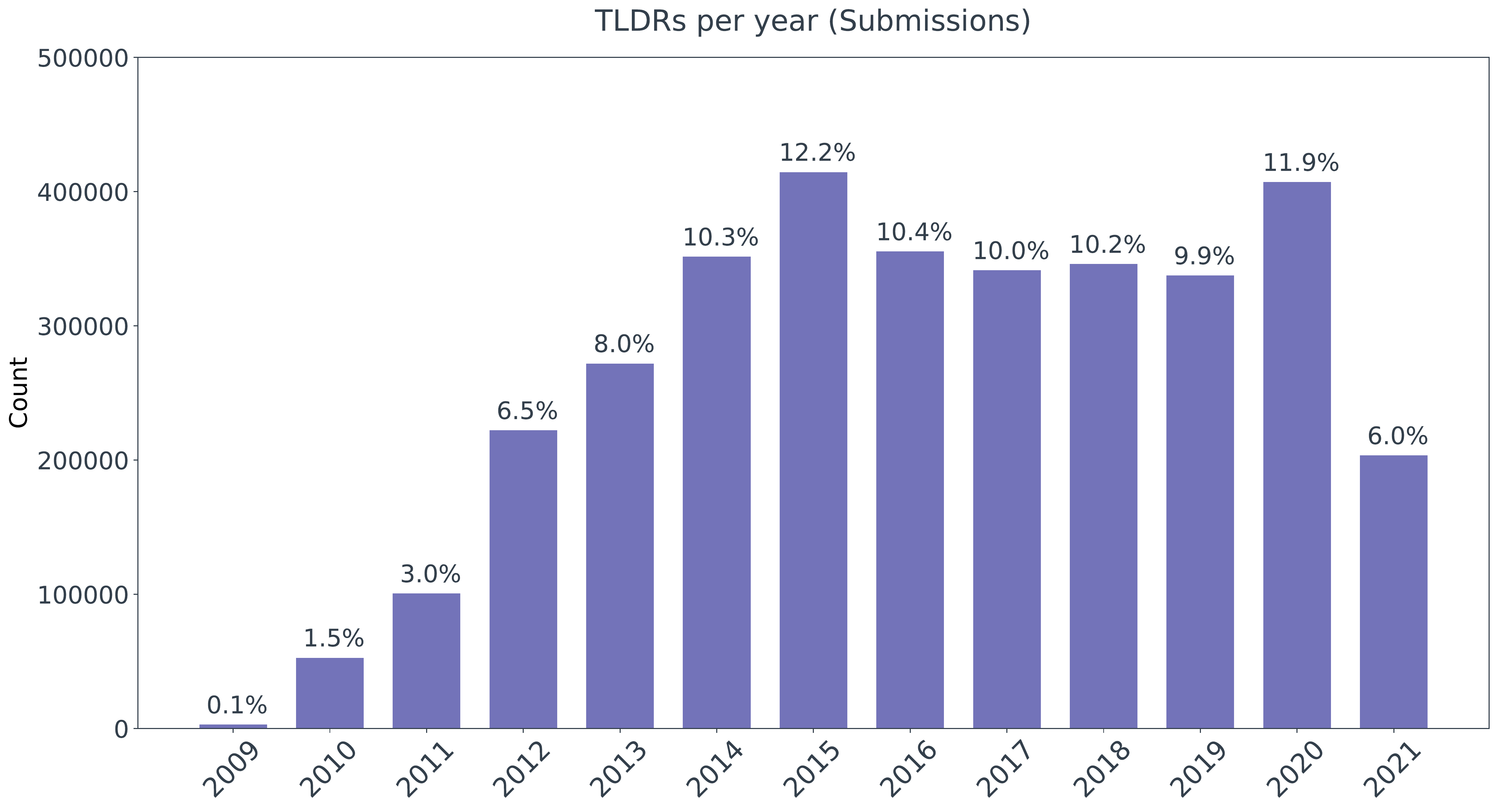}} & \multicolumn{1}{c}{\includegraphics[scale=0.19]{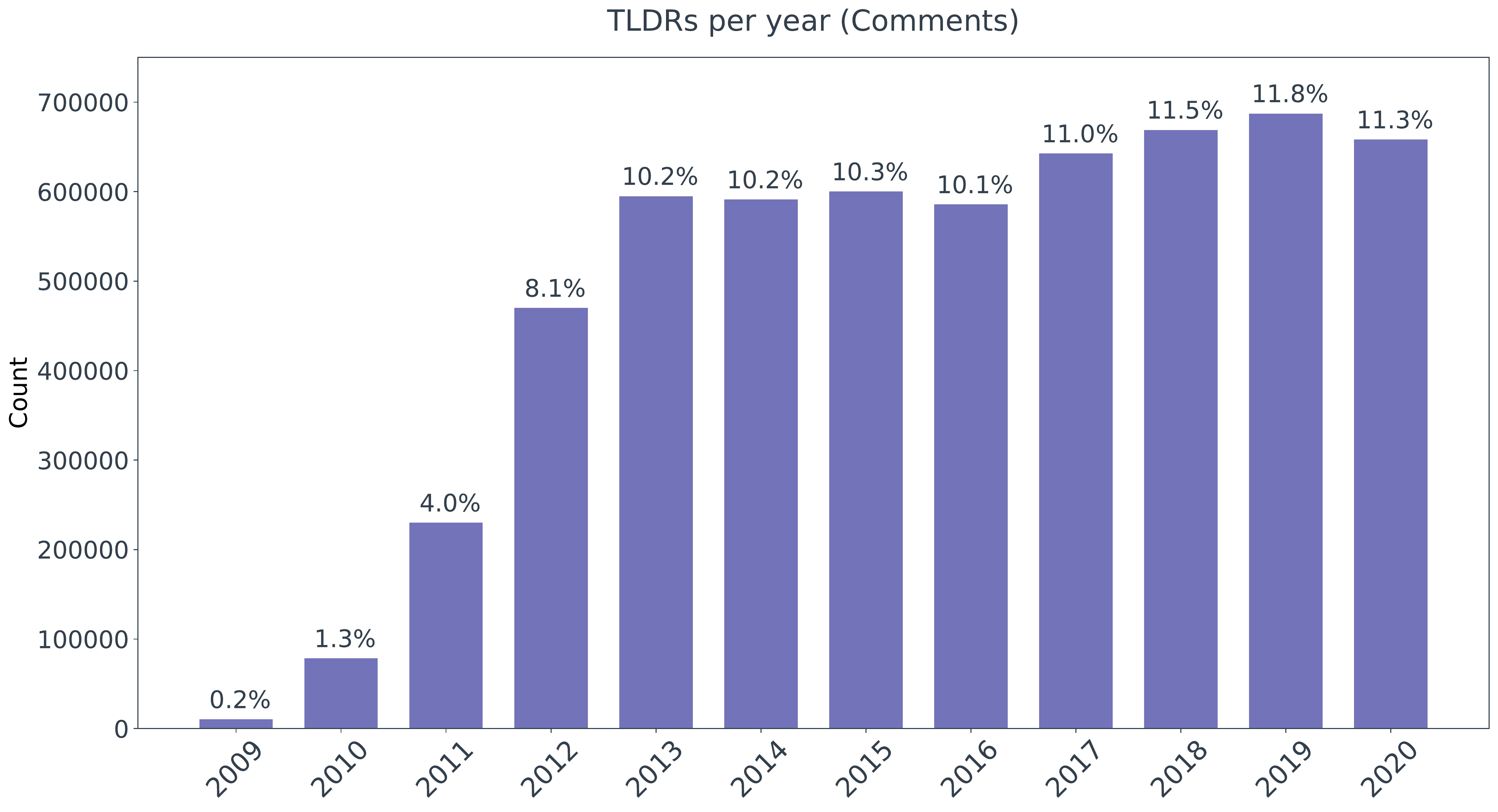}} \\
         \multicolumn{1}{c}{(a) submission-\tldr} & \multicolumn{1}{c}{(b) comment-\tldr} \\
         
        %  \multicolumn{1}{l}{\includegraphics[scale=0.18]{figures/RS2020.pdf}} & \multicolumn{1}{l}{\includegraphics[scale=0.18]{figures/RS2020.pdf}} \\
        %  \multicolumn{1}{c}{(c) submission-TLDR} & \multicolumn{1}{c}{(d) comment-TLDR}\\
        
        %   \\
        %  \includegraphics[scale=0.52]{samples/figs/figanal2.pdf}  \\
        %   \\
         
        %  (b) Sentence graph visualization of our system's generated summary. Green and gray nodes are introductory and non-introductory sentences, respectively. Edge thickness denote the \textsc{Rouge} score strength between pair of sentences.\\
        \label{fig:tldr_proportion}
    \end{tabular}
}
    \caption{The proportion of instances containing \tldr{} in \tldrl{} dataset. As seen, the number of \tldr s is increasing each year. At the time of conducting this research, the submission data dumps are partially uploaded for 2021 (until 2021-06), while there is no comments uploaded for 2021 in the Pushshift repository.
    % At the time of writing this paper, submissions dumps are partially uploaded for 2021 (until 2021-06) while there is no comments uploaded for 2021.
    }
    \label{fig:tldrLFig}
\end{figure*}

Table \ref{tab:ann_score} presents the average \textit{decision score} assigned to the samples on each threshold. The decision score for a given sample is defined as the annotators' average confidence at giving label 1 to that specific sample. If the average confidence score surpasses 0.50, we assign 1 and if it is below 0.50, the sample is annotated with 0. Otherwise, the fifth annotator decides the label. As shown, threshold 0.22 attains the full score in the presence and absence of the context. Overall, this shows that most of the annotators believe the \tldr{} can be distilled considering both oracle sentence and the entire source.

Figure \ref{fig:agreement_rate} shows pair-wise inter-rater $S$ score agreement~\cite{BennetsScore} throughout the annotation process on threshold 0.22, denoting that annotators have mostly slight or fair agreement in labeling process. Specifically, when the context is not provided (i.e., merely with consideration of oracle sentence), raters (2, 4), (2, 3), and (1, 3) have quite a high rate of agreement. On the other hand, most pairs of annotators including (1, 2), (1, 4), and (2, 4) achieve a high agreement rate when the context is given. As the given decision scores ---either only with oracle sentence or provided context--- sum up to 1.0, and considering moderately high agreement rate between the annotators, we decide to sample our \tldrs{} dataset from the instances in that was in threshold 0.22's subset. This leads us to choose human-decided threshold 0.22 as our ground to sample \textbf{H}igh-\textbf{Q}uality \tldr s for constructing \tldrs{} dataset.

 \begin{figure*}[t]
    \centering
    \begin{tabular}{ccc}
        \includegraphics[scale=0.34]{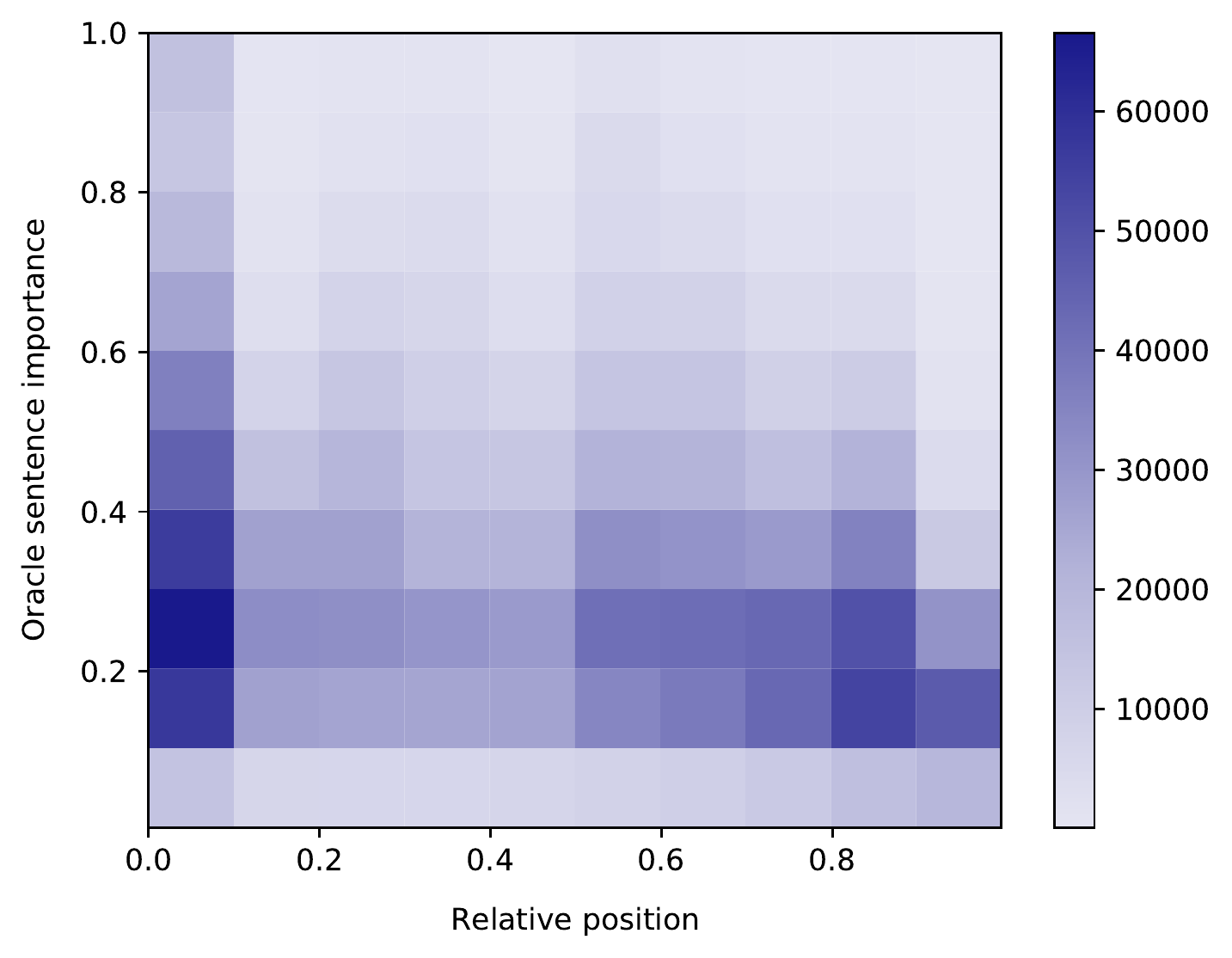} & \includegraphics[scale=0.34]{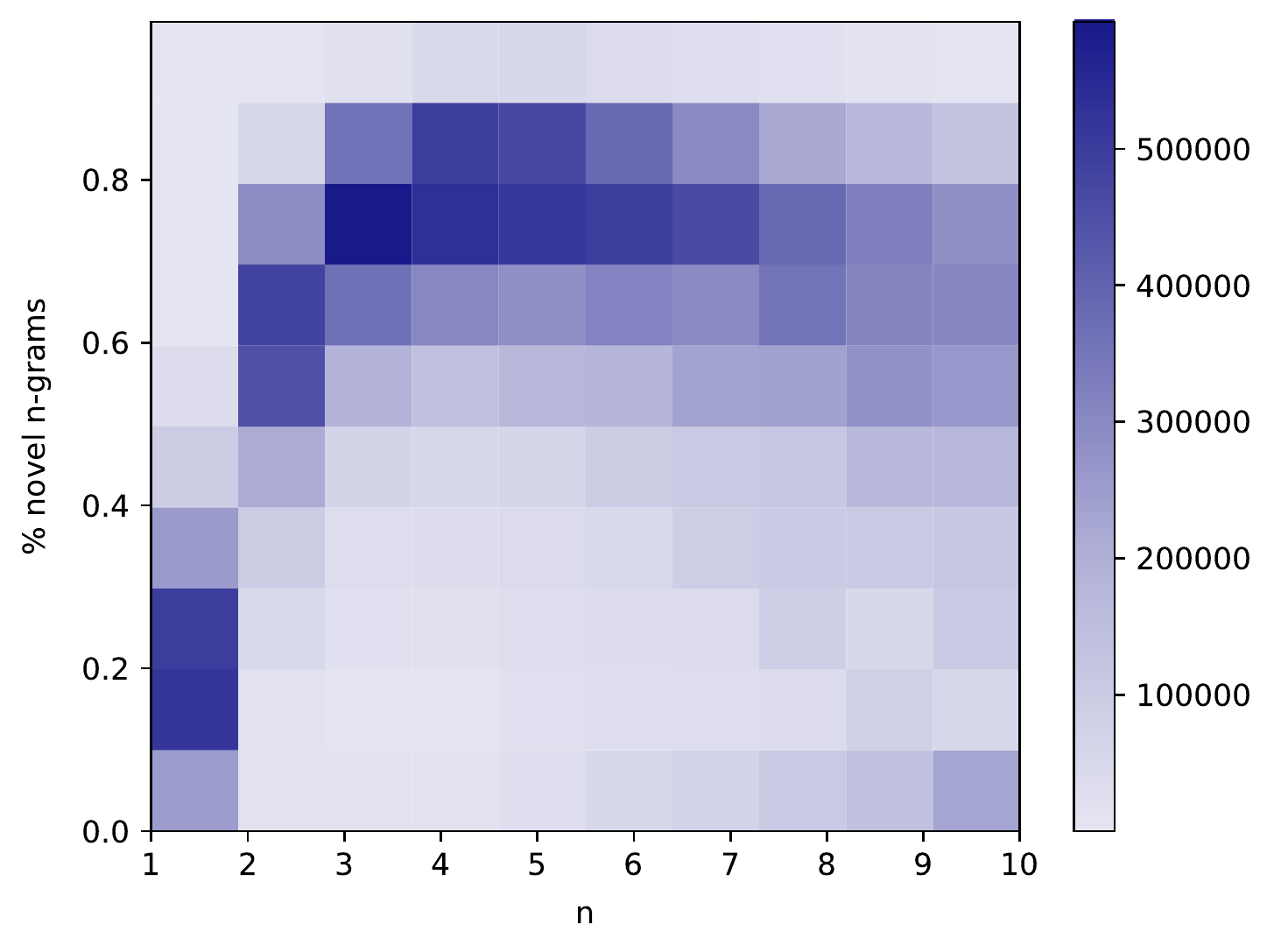} & \includegraphics[scale=0.34]{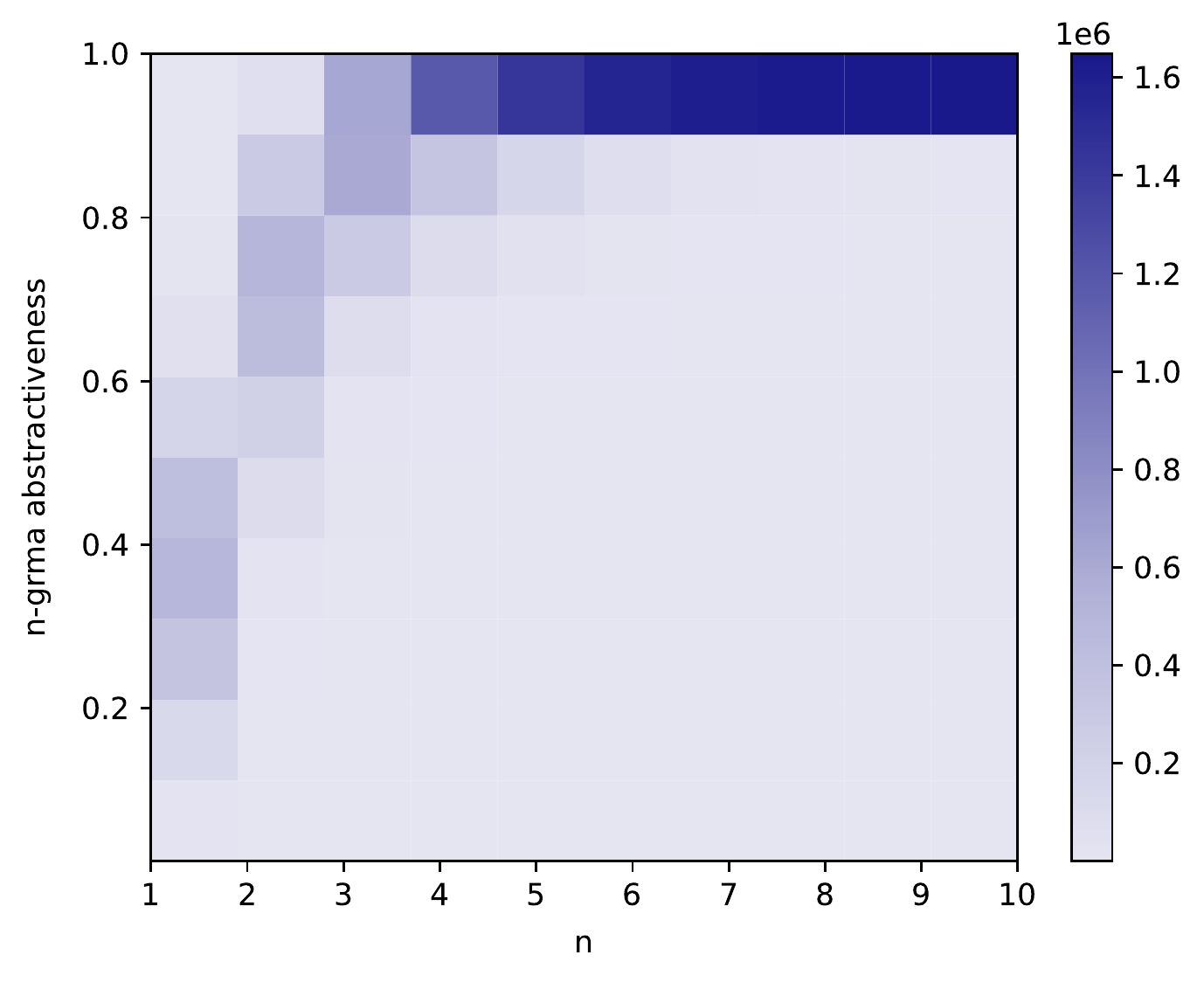} \\
        (a) & (b) & (c) 
    \end{tabular}
    
    \caption{Heatmaps of \tldrs{} showing (a) the oracle sentence's importance to its relative position; (b) percentage of novel n-grams; and (c) n-gram abstractiveness. The heat extent shows the number of the instances within the specific bin.}
    \label{fig:ext-heatmap}
\end{figure*}

%  \begin{align*}\label{eq:pareto mle2}
% \text{Reduced DsName1} & =\{&(p, t) | (p, t) \in \text{DsName1}, \\ & & \text{score}(\text{p},\text{t}) > \lambda \}
% \end{align*}
 
%  \begin{equation}
%      \texttt{Reduced DsName1} = \{(p, t) | (p, t) \in \texttt{DsName1}, score(p,t) > \lambda \}
%  \end{equation}

\subsection{Dataset Analysis}
\label{sec:dataset_analysis}
In this section, we give statistics, along with analyses on the proposed datasets. 

Table \ref{tab:our_ds_stat} shows general statistics of datasets in terms of post and \tldr{} length. As shown, the compression rate~\footnote{Compression rate = $\frac{\text{Avg post length}}{\text{Avg \tldr{} length}}$} is 8.7 and 12.5 in \tldrl{}, and \tldrs{} datasets, respectively. This shows that authors generally tend to write much shorter \tldr s that highly shortens the post's text, which is expected due to the nature of \tldr{} summaries.

\begin{table}[t]
    \centering
    \resizebox{\columnwidth}{!}{%
    \begin{tabular}{lrrrr}
    \toprule
        Dataset & \# instances & post & \tldr{} & Compression \\
        & & (words/sent.) & (words/sent.)& ratio \\ 
         \midrule
         \tldrl & 9,227,437 & 310.3/14.0 & 35.6/2.3 & 8.7  \\
         \tldrs & 1,671,099 &  332.0/15.7 & 27.0/1.8 & 12.5  \\
         \bottomrule
    \end{tabular}
    }
    \caption{Average words length and number of sentences per instance along with the compression ratio in our proposed datasets.}
    \label{tab:our_ds_stat}
    \vspace{-0.4cm}
\end{table}

\begin{table}[t]
    \centering
    % \resizebox{\columnwidth}{!}{%
    \begin{tabular}{lr}
    \toprule
     Dataset Size & 1,671,099 posts \\
     Train Set Size & 1,590,132 posts \\
     \midrule
     Mean Sentence Length & 21.1 tokens\\
     Min Sentence Length & 1 token \\
     Max Sentence Length & 4,370 tokens \\
     \midrule
     Total Vocabulary Size & 1,582,436 \\
     Occurring 10+ Times & 226,754 \\
     Train Vocabulary Size & 1,138,415 \\
     Validation Vocabulary Size & 248,148 \\
     Test Vocabulary Size & 249,079 \\
     
     Train/Test Vocabulary Overlap & 72.5\% \\
     \bottomrule
    \end{tabular}
    % }
    \caption{Detailed statistics of \tldrs{} dataset}
    \label{tab:tldrQ_stat}
\end{table}

Figure \ref{fig:tldrLFig} demonstrates the number of \tldr{} pairs in \tldrl{} across different years. As observed, 83.65\% of these \tldr s occur after 2013 which shows the popularity of this writing style among Reddit users.  We also see a similar trend for years after 2013, each of which constitutes a fixed amount (10\%-12\%) of the dataset. Table \ref{tab:tldrQ_stat} demonstrates the detailed information including data size, sentence length and vocabulary statistics of \tldrs{} dataset. 

As mentioned earlier, we define the oracle sentence to be the one within the longer post that has the highest overlap with \tldr{} summary in terms of \textsc{Rouge-2} and \textsc{Rouge-L} mean scores.  The oracle sentence's relative position in post's text along with its importance is shown in Figure \ref{fig:ext-heatmap} (a). We define the oracle importance score as follows: 
% $\frac{arg)}{\Sigma_i RG(s_i)}$
\begin{equation*}
    \text{oracle importance} = \frac{\max \textsc{RG}_{\textsc{2+L}}(s_i)}{\sum\limits_{s_i \in D}{\textsc{RG}_{\textsc{2+L}}}}
\end{equation*}
% \begin{equation*}
    % s=3
% \end{equation*}
where $D$ is the set of all sentences within the post, and  $s_i$ denotes the $i$th sentence. $\textsc{RG}_{\textsc{2+L}}(.)$ is a function that takes in a post's sentence, and outputs the mean of its \textsc{Rouge-2} and \textsc{Rouge-L} score with respect to \tldr{} summary. Intuitively, the oracle importance score can be framed as the attention score over the oracle sentences when the scoring function is \textsc{Rouge}. Observing Figure \ref{fig:ext-heatmap}, while more of the oracle sentences occur in early parts of the post's text ($\textless 0.10$) with importance score of less than 0.30, it appears that the oracle sentences are spread out across the post's text overall. This observation is substantial, justifying the usability of this dataset for extractive summarization task.

To analyze the abstraction level of \tldrs{} dataset, we plot the percentage of novel n-grams within the \tldr{} summary~\cite{See2017GetTT} in Figure \ref{fig:ext-heatmap} (b), as well as the \tldr 's n-gram abstractiveness~\cite{Gehrmann2019GeneratingAS} in Figure \ref{fig:ext-heatmap} (c) over the all instances in \tldrs{} dataset. As indicated, there are quite a large proportion of novel n-gram words appeared in the \tldr{} summary as the heat extent is mostly concentrated in the upper half of the y-axis. These plots show the promising capability and challenges of this dataset to be used for abstractive summarization models.

% To be more clear, \red{Figure \ref{}} shows the usage rate of TLDRs by the users over the years~\footnote{Usage rate is computed as the proportion of TLDR posts over the entire posts within a given year.}. It is observable that Reddit users have shown more tendency at using TLDRs over the recent years, and this style of writing have become more popular among the Reddit users recently.

\section{Experimental Setup}
\subsection{Baselines}
We benchmark several extractive and abstractive summarization baselines over our two proposed datasets. 

\noindent \textbf{\textsc{BertSumExt.} }\cite{Liu2019TextSW} BertSumExt model is the extractive variant of \textsc{BertSum} which is the \textsc{Bert} Model fine-tuned on text summarization task. In this regard, \textsc{Bert} \texttt{[CLS]} tokens are appended to the start of each input sentence, and their associated representations are used to predict if the sentence should be included in the final summary or not. 

\noindent \textbf{\textsc{BertSumAbs.} }\cite{Lewis2020BARTDS} \textsc{BertSumAbs} is the abstractive model of \textsc{BertSum}, where a Transformers-based decoder is added to the \textsc{Bert} Encoder. 

% \noindent \textbf{\textsc{BertSumExtAbs.} }\cite{Liu2019TextSW} Another variant of \textsc{BertSum} that combines extractive and abstractive models to benefit from the shared knowledge that between these two tasks. Specifically, a \textsc{BertSumExt} model is first trained to identify salient sentences, and then is used along with a Transformers-based decoder to produce abstractive summaries. 

\begin{table*}[t]
\centering 
\begin{center}
\begin{tabular*}{\textwidth}{l@{\hspace{1.5em}}l@{\hspace{0.7em}}l@{\hspace{0.7em}}l@{\hspace{0.7em}}l@{\hspace{1em}}l@{\hspace{0.7em}}l@{\hspace{0.7em}}l@{\hspace{0.7em}}l}
\toprule
      & \multicolumn{3}{c}{\tldrl} &   & \multicolumn{3}{c}{\tldrs} \\
 \cline{2-4}  \cline{6-8}
 Model                    & \small RG-1(\%)  &\small RG-2(\%)  &\small RG-L(\%)  & &\small RG-1(\%)  &\small RG-2(\%)  &\small RG-L(\%) \\ 

 \midrule

   \textsc{BertSumExt}~\cite{Liu2019TextSW}                   &  {20.94} &	 {4.98} &	 {14.48}  & &  {28.40}&	 {11.35}&	21.38 \\

  \textsc{BertSumAbs}~\cite{Liu2019TextSW}                   &  {23.05} &	 {9.48}&	 {18.07}  & &  {28.96}&	 {12.08}&	 {22.08} \\

%  \textsc{BertSumExtAbs}~\cite{Liu2019TextSW}                   &  {48.42} &	 {19.71} &	 {21.47}  & &  {48.82}&	 {20.89}&	 {23.37} \\

  \textsc{Bart}~\cite{Lewis2020BARTDS}                   &  {23.59} &	 {9.69} &	 {18.62}  & &  {32.44}&	 {14.85}&	 {27.39} \vspace{0.3em}
          \\\hdashline \vspace{-0.3cm}\\
  \textsc{Oracle-Ext}                   &  {30.26} &	 {9.74} &	 {20.60}  & &  {45.29}&	 {25.47}&	 {36.86} \\

%   \midrule
\bottomrule
\end{tabular*}
\end{center}
\caption{\textsc{Rouge (F1)} results of the state-of-the-art summarization models on the test sets of the proposed \tldr{} summarization datasets (\tldrl, and \tldrs).}
\label{tab:arxsum}
\label{tab:main}

\end{table*}

\noindent \textbf{\textsc{Bart.} }\cite{Lewis2020BARTDS} \textsc{Bart} is a regressive autoencoder model that is pre-trained by first corrupting the text with an arbitrary noising function, and secondly, trying to reconstruct the original input text. \textsc{Bart} is particularly effective when fine-tuned on text generation tasks such as summarization. As \textsc{Bart} has both encoder and decoder pre-trained, it can be perceived as an extension to general \textsc{Bert} models in which only encoder is pre-trained. 

\subsection{Dataset}
We randomly split our datasets to construct training, validation, and test sets. Specifically, for \tldrl , we use 99-0.5-0.5 split which results in 9,139,935 (train), 43,753 (validation), and 43,749 (test) instances. To split \tldrs , we use 95-2.5-2.5 division yielding 1,590,132 (train), 40,481 (validation), and 40,486 (test) pairs.

\subsection{Training and Hyper-parameters}
To train the summarization models, we utilize HuggingFace's Transformers~\cite{Wolf2020Transformers} for \textsc{Bart}, and the open implementation~\footnote{\url{https://github.com/nlpyang/PreSumm}} of \textsc{BertSumExt}, \textsc{BertSumAbs}. We use warm-up steps of 32K, and 20K for \textsc{Bart} and \textsc{BertSum} variants, respectively. The AdamW optimizer~\cite{Loshchilov2019DecoupledWD} is used with learning rate of $3e-5$, beta parameter of $0.98$, and weight decay of $0.01$ for \textsc{Bart} model. For \textsc{BertSum} variants, we use the default Adam~\cite{Kingma2015AdamAM} optimizer with learning rates of $2e-3$ for the encoder, and $1e-2$ for the decoder as suggested by the main paper~\cite{Liu2019TextSW}. For all models, we use cross-entropy loss function. We train the models on 8 Nvidia Tesla V100 GPUs for 5 epochs with early stopping of the training when the validation loss does not decrease for 3 consecutive validation steps. The validation step is done every 25K training steps. To visualize and keep track of the learning process, we use Weight and Biases~\cite{wandb} toolkit.

\section{Experimental Results}
Table \ref{tab:main} presents the performance of the state-of-the-art summarization models on our proposed datasets in terms of \textsc{Rouge-1}, \textsc{Rouge-2}, and \textsc{Rouge-L} scores. As indicated, \textsc{Bart} outperforms all other models across all \textsc{Rouge} variants in both datasets. This is expected as \textsc{Bart}'s both encoder and decoder have been pre-trained on a large amount of unlabelled data, unlike \textsc{BertSum} variants that only have pre-trained encoders. 

Comparing abstractive models with \textsc{BertSumExt}, we observe relatively large performance gap. This might be due to the fact that \tldr s in both \tldrl{} and \tldrs{} datasets are rather abstractive than extractive as also shown in Section \ref{sec:dataset_analysis}. Yet with the existence of such a huge gap, the \textsc{Oracle-Ext} (i.e., upper bound of an extractive summarizer) scores prove that more developed extractive summarizers can perform out-of-the-box and mitigate this gap. The performance gap on \tldrl{} brings various challenges to develop summarization models that better fit on the larger dataset that include noisy data~\cite{Kumar2020NoisyTD}. This noise might be handled via methods such as noise-aware training models~\cite{Namysl2020NATNT}, while enabling the models to benefit from the large-scale \tldrl{} dataset. We leave this part for future work. It has to be mentioned that automatic evaluation of summarization continues to be an issue and while this dataset does not solve that, instead can be used with any evaluation metric as they evolve.
\begin{figure}[h]
    \centering
    
    \begin{tabular}{l}
         \includegraphics[scale=0.38]{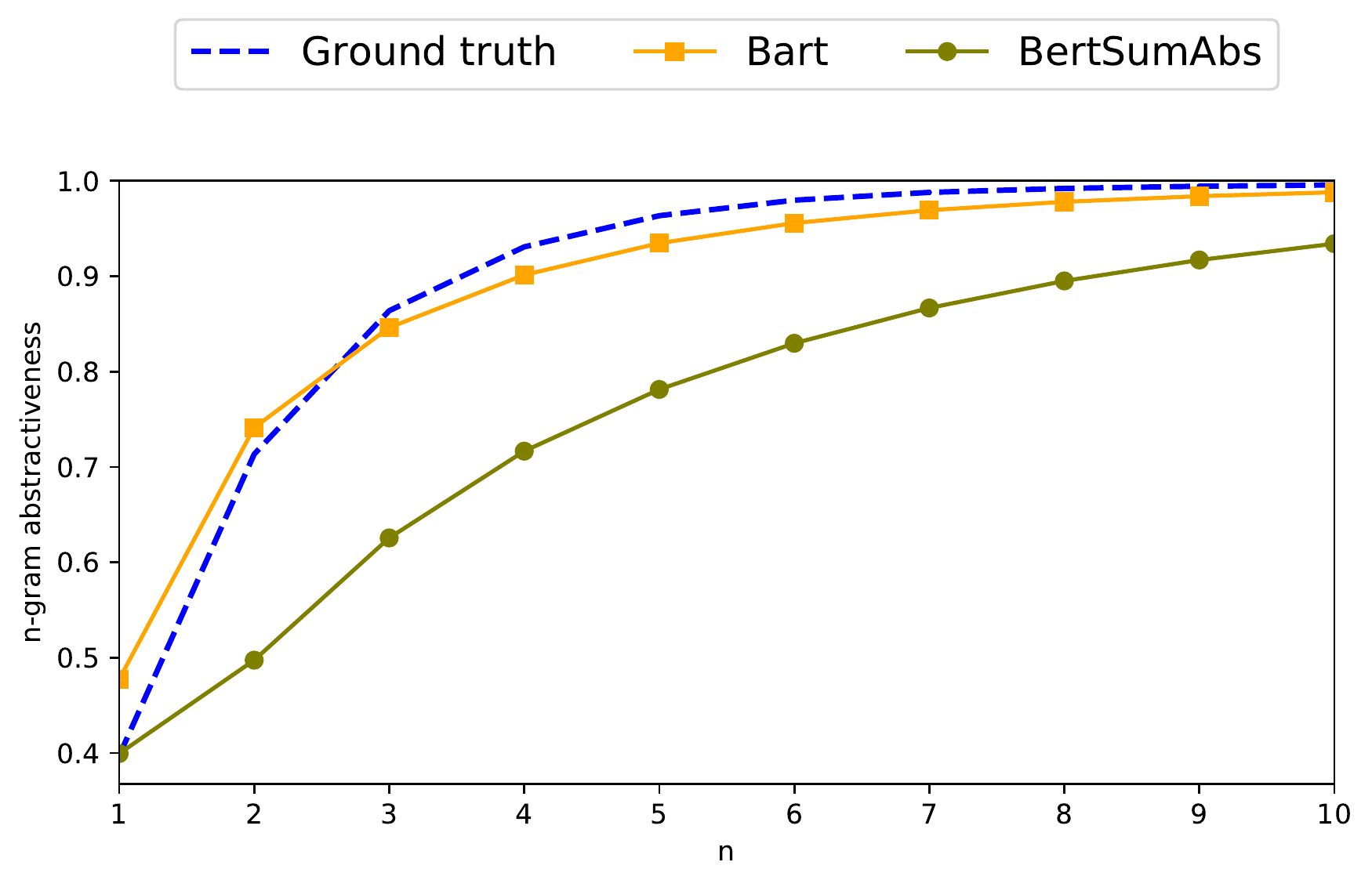}
         \\
         \includegraphics[scale=0.38]{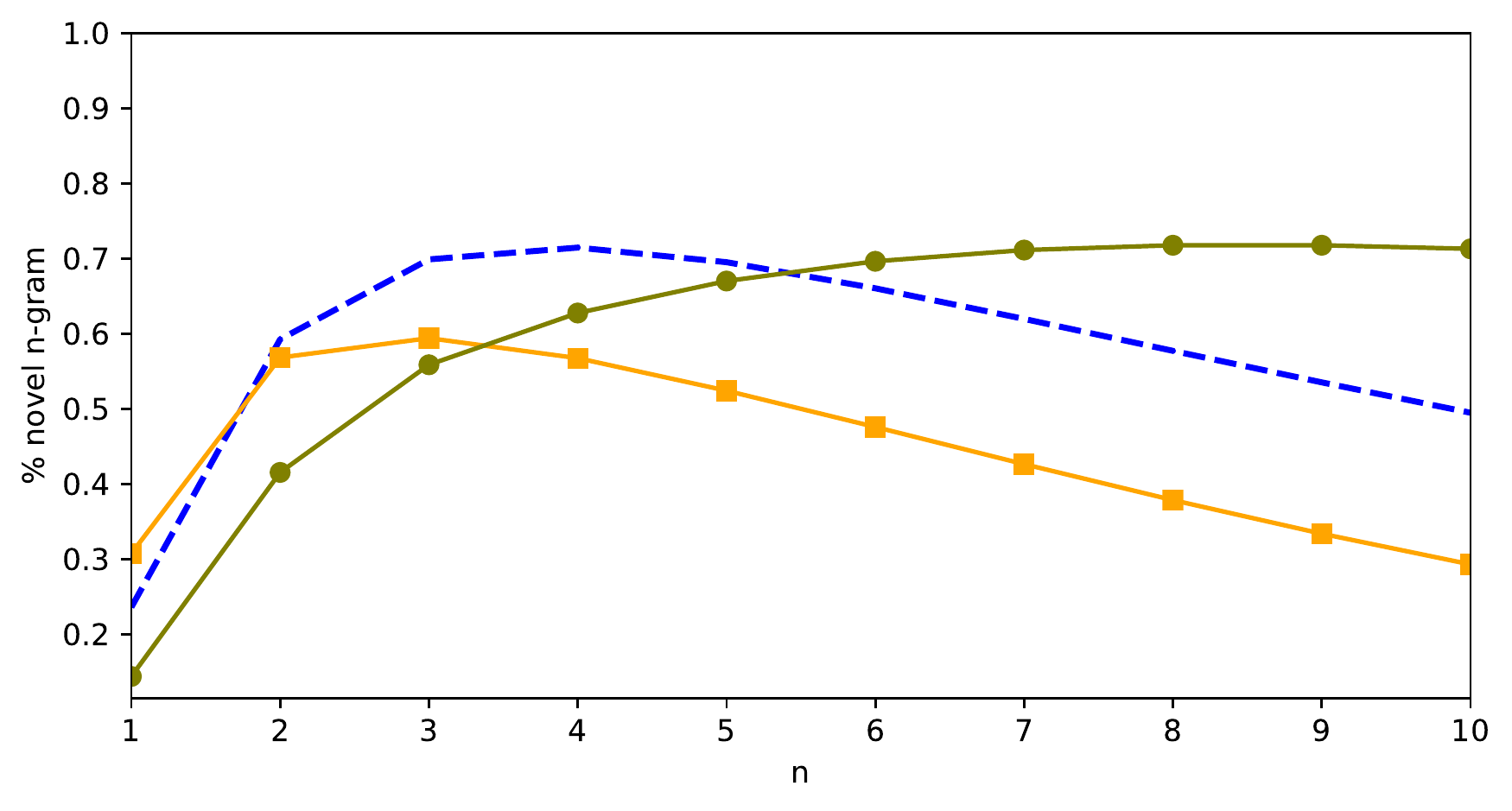}
    \end{tabular}
    
    \caption{The n-gram abstractiveness and percentage of novel n-gram metrics across different n-grams on \tldrs 's test set. As seen, \textsc{Bart} generates more abstractive summaries than \textsc{BertSumAbs} as it mitigates the gap between \textsc{BertSumAbs} and ground truth summary.}
    \label{fig:abstractiveness}
\end{figure}

\begin{figure}[h]
    \centering
    \includegraphics[scale=0.65]{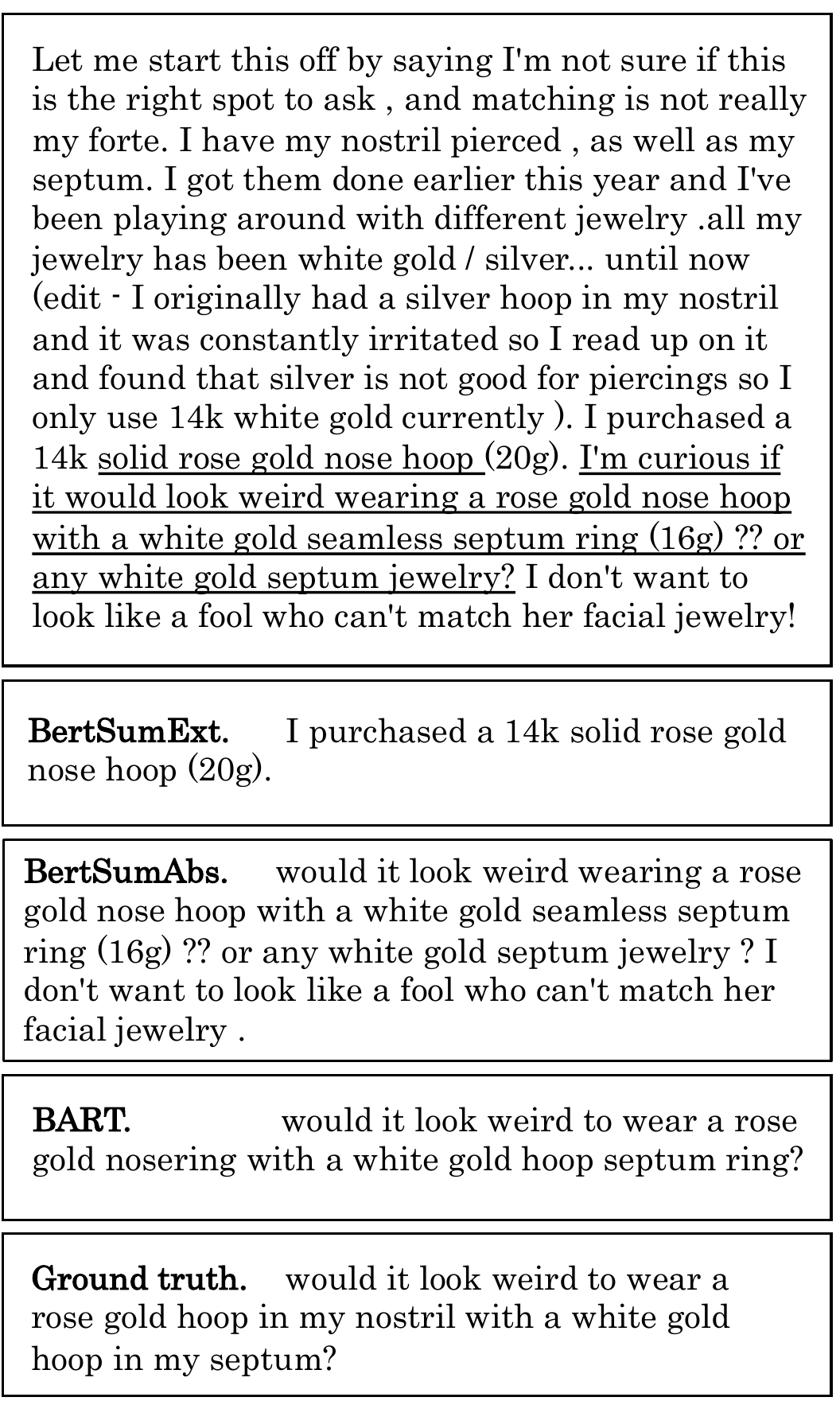}
    \caption{A sample from \tldrs{} test set along with the model generated summaries. Underlined text in source shows the important regions of the source for generating \tldr{} summary. }
    \label{fig:gen_ex}
\end{figure}

\section{Analysis}
To gain insights into the qualities of summarization model, we analyze the outputs generated by the models. The diagrams demonstrating n-gram abstractiveness and percentage of novel n-grams, generated by \textsc{Bart} and \textsc{BertSumAbs}, are plotted in Figure \ref{fig:abstractiveness}. As observed, \textsc{Bart} model appears to have a similar trend to the ground truth \tldr s. On the other hand, \textsc{BertSumAbs} model has increasing n-gram abstractiveness, and novel n-gram percentage with increasing n. It is also interesting that after 6-gram, \textsc{BertSumExt} model reaches a plateau when generating novel n-grams, but we a drop after 3-grams for \textsc{Bart} and the ground truth \tldr s. This shows that from 1-gram to 3-gram, there are increasing number of novel words appeared in the ground-truth and \textsc{Bart}, but after that, they both tend to copy n-grams rather than generating those.

To understand the limitation and qualities of current state-of-the-art summarization models, we conduct a qualitative analysis on several samples from \tldrs{} dataset, of which one is shown in Figure \ref{fig:gen_ex}. Analyzing this sample, we observe that \textsc{Bart} generated a better summary in terms of faithfulness to the ground truth \tldr . On the other hand, while \textsc{BertSumAbs} could identify the important region of the source document, it has produced a longer \tldr{} with additional information that is present in the source, but not in the ground truth \tldr{} summary. \textsc{BertSumExt} model could have identified a source sentence that is partly in connection with the ground truth \tldr, but it leaves out the most important sentence as the oracle to be extracted. Considering the upper performance of extractive summarizers (i.e., \textsc{Oracle-Ext} score in Table \ref{tab:main}), we believe that there is a large room for improvement on this dataset. Investigations of more advanced models remain for future work. 
\section{Conclusion}

In this paper, we proposed two large-scale summarization datasets called \tldrl, and \tldrs. The \tldrl{} dataset contains over 9 millions Reddit post-\tldr{} instances. To distill a more fine-grained dataset out of \tldrl{}, we sample high-quality instances with the help of human annotations to construct \tldrs . Our analyses over \tldrl{} and \tldrs{} datasets show its usability for performing both extractive and abstractive summarization tasks. We further establish extractive and abstractive baseline results using state-of-the-art summarization models on both datasets. We hope our datasets can pave the path for future studies in this direction. 

\section*{Acknowledgements}
We warmly thank the anonymous reviewers as well as Tracy King for their helpful feedback and suggestions.

% Text 

% \section{Analyses}

% \begin{itemize}
% \item Plot/statistics/tables ideas:
% \item TLDR per year or month
% \item TLDR per year or month, for different filtering rules (do people use TLDR less carefully over time?)
% \item Presences of TLDR vs. upvotes 
% \item Breakdown of nb of TLDR between submissions and comments. See nb of comments and submissions removed by the filters  
% \item TLDR per sub (we can list e.g. the top 10/15/20 with the most TLDR), normalized per nb of posts
% \item model performances on training/dev/set, when using different filter rules
% \item different filter rules vs. human annotations
% \item  we should probably have some tables in the paper to give some stats on comments/submissions length and TLDR length
% \item  ng to compare the performance of the summarizers on comments only vs. submission only vs. two combined
% \item compare karma and length of post with a TLDR vs. (do 1 boxplot for all posts with a TLDR vs. 1 boxplot for all posts without a TLDR)
% \item Table of the most frequently used TLDR keywords \url{https://adoberesearch.slack.com/archives/D02368L163D/p1627954754094100}
% \item another cool graph we could do for the paper: \url{https://aclanthology.org/N19-1168.pdf}  fig. 4
% \item plot showing train set size vs. convergence speed
% \item plot showing train set size vs. performance on the test set
% \end{itemize}

% \section{Paper to cite}
% \begin{itemize}
% \item  https://arxiv.org/abs/2001.08435 The Pushshift Reddit Dataset
% \item  Fei Liu papers
% \end{itemize}

% \section*{Acknowledgements}

% This document has been adapted

\bibliography{anthology,custom}
\bibliographystyle{acl_natbib}

\appendix

% \section{Example Appendix}
% \label{sec:appendix}

% This is an appendix.

\end{document}